\documentclass[sigconf]{acmart}
\usepackage{graphicx}
\usepackage{hyperref}
\usepackage{listings}
\usepackage{booktabs}
\usepackage{multirow}
\usepackage{amsmath}
\usepackage[most]{tcolorbox}
\usepackage{xcolor}
\usepackage{booktabs} 
\usepackage{amsmath}
\usepackage{amsthm}
\usepackage{multirow,tabularx,ragged2e}
\usepackage{adjustbox}
\usepackage{graphicx}
\usepackage{hyperref}
\usepackage{enumitem}
\usepackage{lipsum}
\usepackage{multirow}

\copyrightyear{2024}
\acmYear{2024}
\setcopyright{rightsretained}
\acmConference[GeoAnomalies'24]{1st ACM SIGSPATIAL International Workshop on Geospatial Anomaly Detection}{October 29-November 1, 2024}{Atlanta, GA, USA}
\acmBooktitle{1st ACM SIGSPATIAL International Workshop on Geospatial Anomaly Detection (GeoAnomalies'24), October 29-November 1, 2024, Atlanta, GA, USA}
\acmDOI{10.1145/3681765.3698463}
\acmISBN{979-8-4007-1144-2/24/10}

 \def\andreas#1{{\footnotesize{\sc\textcolor{red}{}}}}
 \def\andi#1{{\footnotesize{\sc\textcolor{red}{}}}}
 \def\hossein#1{{\footnotesize{\sc\textcolor{blue}{}}}}
 \def\lance#1{{\footnotesize{\sc\textcolor{blue}{}}}}

\renewcommand\footnotetextcopyrightpermission[1]{}

\begin{document}

\title{Neural Collaborative Filtering to Detect Anomalies in Human Semantic Trajectories}
\author{Yueyang Liu}
\orcid{}
\affiliation{%
\institution{Emory University, Atlanta, USA}
  \city{}
  \state{}
  \country{}
}
\email{yueyang.liu@emory.edu}

\author{Lance Kennedy}
\orcid{0009-0004-6815-2219}
\affiliation{%
  \institution{Emory University, Atlanta, USA}
  \city{}
  \state{}
  \country{}
}
\email{lance.kennedy@emory.edu}
\author{Hossein Amiri}
\orcid{0000-0003-0926-7679}
\affiliation{%
  \institution{Emory University, Atlanta, USA}
  \city{}
  \state{}
  \country{}
}
\email{hossein.amiri@emory.edu}
\author{Andreas Z{\"u}fle}
\orcid{0000-0001-7001-4123}
\affiliation{%
  \institution{Emory University, Atlanta, USA}
  \city{}
  \state{}
  \country{}
}
\email{azufle@emory.edu}

\renewcommand{\shortauthors}{Liu et al.}
\renewcommand{\shortauthors}{Liu et al.}
\begin{abstract}
    Human trajectory anomaly detection is critical for applications such as security surveillance and public health, yet most existing methods focus on vehicle-level traffic, with limited attention to human-level trajectories. Due to the inherent sparsity of human trajectory data, machine learning approaches are favored for detecting complex patterns. However, concerns about model biases and robustness have highlighted the need for more transparent and explainable solutions.
In this paper, we propose a lightweight anomaly detection model specifically designed to detect anomalies in human trajectories. We propose a Neural Collaborative Filtering approach to model and predict normal mobility.
Our method is designed to model users' daily patterns of life without requiring prior knowledge, thereby enhancing performance in scenarios where data is sparse or incomplete, such as in cold start situations. 
Our algorithm consists of two main modules.  The first is the collaborative filtering module, which applies collaborative filtering to model normal mobility of individual humans to places of interest. The second is the neural module, responsible for interpreting the complex spatio-temporal relationships inherent in human trajectory data. To validate our approach, we conducted extensive experiments using simulated and real-world datasets comparing to numerous state-of-the-art trajectory anomaly detection approaches.
\vspace{-0.3cm}
\end{abstract}
\vspace{-0.5cm}
\vspace{-0.5cm}

\maketitle
\section{Introduction}
\label{sec:introduction}
\begin{figure}[t]
    \centering
    \includegraphics[width=0.950\linewidth,height=5cm,trim=0cm 0cm 0cm 0cm,clip]{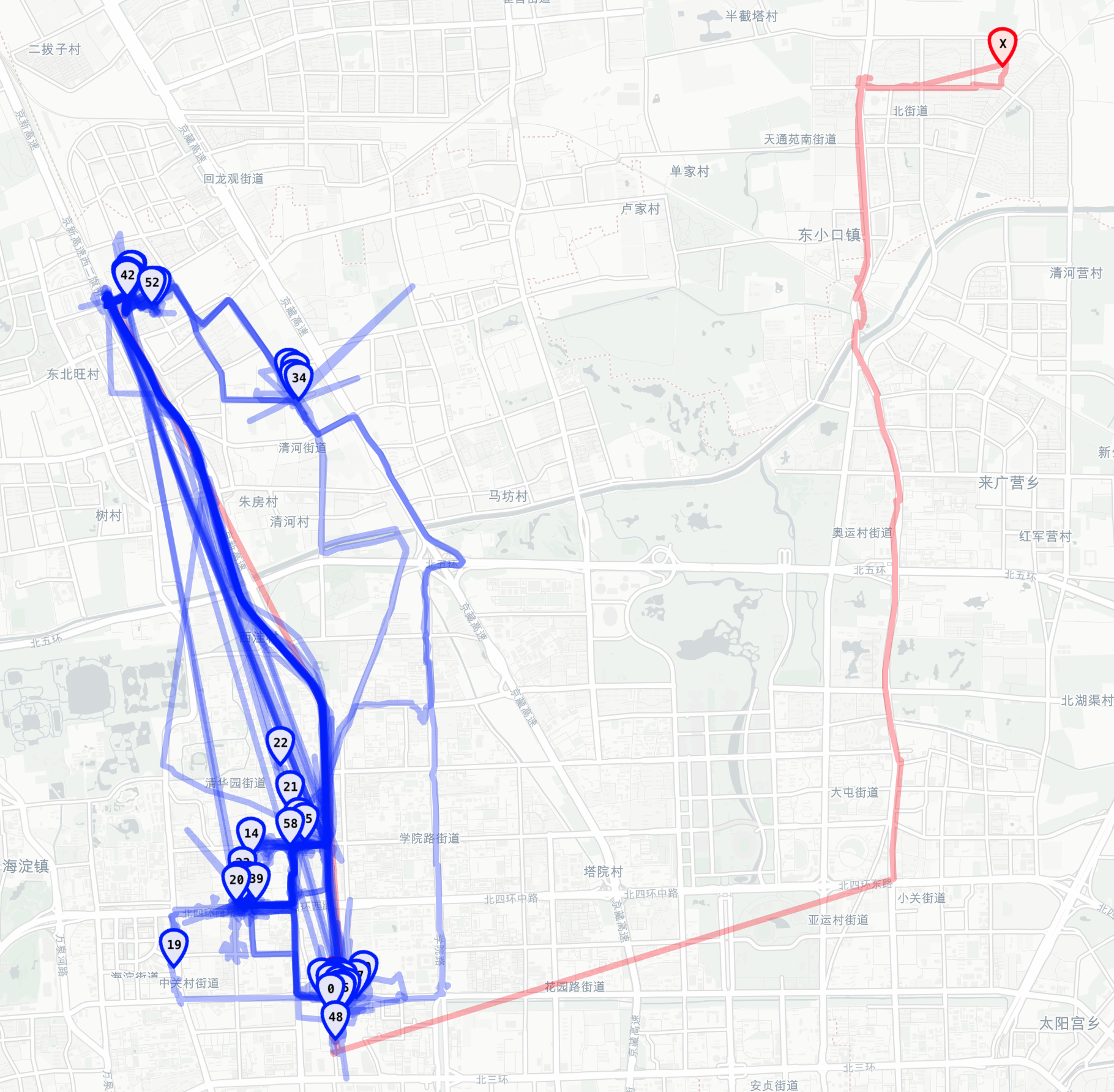}
    \vspace{-0.4cm}
    \captionsetup{width=1\linewidth}
    \caption{A sample of anomaly trajectory (red) and Normal trajectory (blue).\vspace{-0.3cm}}
    \label{fig:sp_outlier}
    \vspace{-0.3cm}
\end{figure}


A trajectory is a trace generated by a moving object in a geographic space, usually represented by a sequence of chronologically ordered points~\cite{zheng2015trajectory}. 
A semantic trajectory~\cite{parent2013semantic} is a sequence of chronologically ordered points where some points may be enriched with semantic information such as the type of a visited place of interest. Figure~\ref{fig:sp_outlier} shows an example of a semantic trajectory of a user in the GeoLife~\cite{geolife} dataset in Beijing, China. Blue points form lines to visualize the trajectory of the user and location markers highlight staypoints that can be reverse geocoded to obtain semantic information of visited places.

An anomaly is commonly defined as a data point that is significantly different from the remaining data. Anomalies are also referred to as abnormalities, discordants, deviants, or anomalies in the data mining and statistics literature~\cite{aggarwal2017introduction,stanford2024numosim,duan2024bayesics}.
In this example, the monitored user mainly stays within a certain area and only has a single long-distance trip.
Finding anomalous behavior in semantic trajectories has many applications:
\begin{itemize}[noitemsep,
        nosep,
        leftmargin=5pt,
        labelsep=0pt,
        itemindent=0pt]
    \item {\bf Infectious Disease Early-Warning:} a large number of people staying at home from work may indicate an outbreak of an infectious disease even before people notice more severe symptoms~\cite{ostfeld2005spatial}.
    \item {\bf Elder Monitoring:} A person suffering from dementia wandering the park at 11pm may indicate that the person might be lost and in need of assistance~\cite{stavropoulos2020iot,tolea2016trajectory}.
    \item {\bf Child Safety:} A child being taken to a far away house when they should be at school may indicate a possible child abduction~\cite{d2017design}. 
\end{itemize}
Despite these important applications, the implementation of systems to support them is challenging, as it becomes very difficult to decide: What part of a trajectory is anomalous? What part is normal?
For example, Figure~\ref{fig:sp_outlier} shows the six-week trajectory of a user in GeoLife. This trajectory allows us to observe the users patterns of life: including their home location, their work or school location, and other places the user likes to visit. Existing work on trajectory-based user identification has shown that an individual's patterns of life are like a unique fingerprint~\cite{de2013unique,seglem2017privacy}, thus allowing identification of any individual human even among a large crowd of users. 
But the goal of this work is not to identify users, but to detect anomalous behavior within a user's trajectory. Such anomalies, of having users deviate from their normal patterns of life, may indicate a stolen phone or indicate distress, such as an abducted child or a lost elder with dementia.
To illustrate the problem of trajectory anomaly detection, Figure~\ref{fig:sp_outlier} depicts the multi-week trajectory of GeoLife user 15. Let us assume that this trajectory belongs to 
a nine-year-old child named Maria. Also, assume that the one trip to the East of the City happened during a time when Maria would normally be at school. Do we think that this trip is anomalous? Could it be that Maria was abducted? We don't really have enough information to answer this question without knowing the semantics of the visited place. Maybe this is the location of the National History Museum that high school classes commonly visit and Maria is happily looking at dinosaur fossils? Maybe this is the location of the home of Maria's friend, whom other children like Maria often visit to skip school. None of these cases would be cause for safety concerns. We use this stylized example to show that we need to know more than just time and location to infer anomalous behavior: We need the semantics of places: What other individuals visited this place? Even though Maria never visited this place: Is it normal for others like Maria to visit this place?

Over the past few decades, a considerable amount of research has been dedicated to trajectory anomaly detection across multiple domains, including human mobility~\cite{meng2019overview,belhadi2020trajectory}, maritime~\cite{machado2019vessel}, and transportation~\cite{CuiKinematicVehicle}. These works mainly define anomaly trajectories as those that exhibit a large distance (in space and time) from other trajectories. However, the above example shows that space and time alone do not suffice to discriminate between normal-but-not-yet-observed and anomalous movement. To fill this research gap, we propose to consider using the semantics of locations: What is the (latent) semantic or purpose of a visit?

But how can we infer the latent semantics of a trip when trajectory data only includes location information? To answer this question, we propose to use a Collaborative Filtering (CF) approach. 
CF is traditionally used in recommendation systems and leverages the known preferences of a group of users to make recommendations or predict the unknown preferences of other users~\cite{suncfsurvey}. For example, in the Netflix Prize competition, the winning proposal, "Cinematch," demonstrated the effectiveness of CF by surpassing Netflix's own system by more than 6\% ~\cite{bennett2007netflix}.
The advantage of CF is that it does not require semantic data to infer semantic information: In the movie recommendation example, without needing to know the semantics of movies (genres, actors, etc.) and without knowing the preferences of users (children, action, anime, love stories), CF allows one to estimate the rating that a user will give to a movie by looking at similar users (similar in terms of what movies they liked) and similar movies (similar in terms of liked by similar users).

While these systems estimate the likelihood that a user will like a movie, our idea is to use collaborative filtering to estimate the likelihood that a person will visit a place. Even though Maria never visited the National History Museum and even though the Museum is far from her home and school, our model may give a high likelihood for Maria to visit the Museum. That's because other children (who visit places similar to Maria, such as elementary schools, playgrounds but not universities and bars) have also visited the Museum in the past. 
The CF will give us, for each (person, place)-pair a likelihood that the user would (normally) visit this place. For (person, place) pairs for which the model estimates a very low likelihood of a visit, the model tells us: ``This person should not normally visit this place''. If we observe such a visit in the data, the model will be surprised. We can measure this surprise as the difference between the expected value and the observed value in test and use it report anomalous.

\begin{figure}[t]
    \centering
    \includegraphics[width=0.959\linewidth]{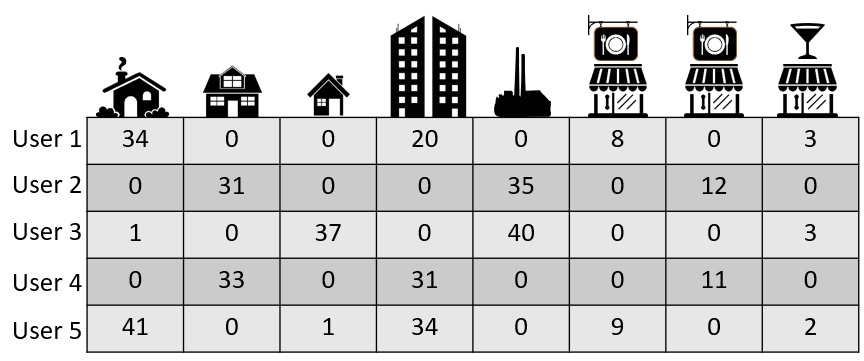}
    \vspace{-0.4cm}
    \captionsetup{width=1.\linewidth}
    \caption{A stylized example of a User-POI matrix \andi{need to move caption text to intro} \vspace{-0.2cm}}
    \label{fig:upmatrix}
    \vspace{-0.2cm}
\end{figure}
\begin{figure}[t]
    \centering
    \includegraphics[width=0.959\linewidth]{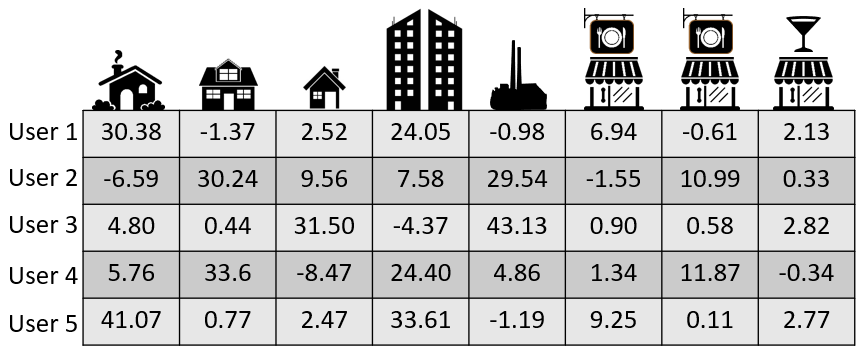}
    \vspace{-0.4cm}
    \captionsetup{width=0.99\linewidth}
    \caption{Expected User-POI visits obtained from the User-POI Matrix of Figure~\ref{fig:upmatrix}. Obtained via Randomized Singular Value Decomposition having three latent components. \vspace{-0.4cm}}
    \label{fig:svd_upmatrix}
    \vspace{-0.2cm}
\end{figure}

\begin{figure}[t]
    \centering
    \includegraphics[trim = 0.1cm 0.1cm 0.1cm 0.1cm , clip,width=0.90\linewidth]{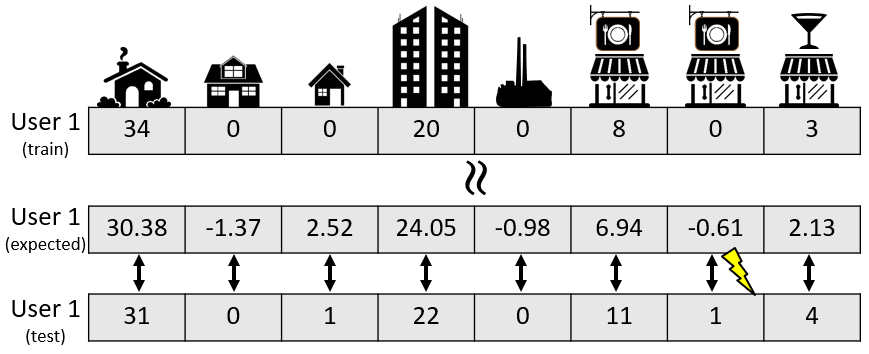}
    \vspace{-0.4cm}
    \captionsetup{width=0.99\linewidth}
 \caption{Measuring Surprise between the expected User-POI Matrix and observed User-POI visits.\vspace{-0.5cm} }
    \label{fig:res_upmatrix}
    \vspace{-0.35cm}
\end{figure}

To illustrate this concept, Figure ~\ref{fig:upmatrix} presents a stylized example of a User-POI matrix. This matrix captures the number of visits by each of five users to eight different Points of Interest (POIs). For instance, User 1 visits a house (likely their home) 34 times, an office building (likely their workplace) 20 times, a restaurant (possibly their favorite) 8 times, and a bar three times. User 2 visits entirely different locations, while User 5 lives in the same area as User 1 and frequents similar places.
Figure ~\ref{fig:svd_upmatrix} shows the same matrix after factorization using Singular Value Decomposition (SVD) with three latent components and re-expansion of the factor matrices. Thus, Figure ~\ref{fig:svd_upmatrix} is the result of a coder-decoder approach: SVD encodes the matrix into a compressed representation that aims at filtering noise and retaining signal. Then, the re-expansion of the factor matrices decodes this latent representation into a full matrix after dropping the least important principal components. Intuitively, we can interpret Figure ~\ref{fig:svd_upmatrix} as a recommendation matrix: Where does the SVD model think each user should normally go?
Figure ~\ref{fig:res_upmatrix} then measures the reconstruction difference between the original User-POI matrix and the SVD-generated matrix. A high surprise score indicates that the user is more likely to visit a POI that they are not typically expected to visit. We see that User 1 visited two places in the test period that were never visited by User 1 in the train period. However, the model judges the visit to the house (third column in Figure ~\ref{fig:res_upmatrix}) as non-surprising. Because others like User 1 visited this house, the model even expected User 1 to visit this house 2.52 times. However, the new restaurant that User 1 visited (second column from the right in Figure ~\ref{fig:res_upmatrix}) yields a high model surprise. The model estimated User 1 to visit this place -0.61 times. This means that the model did not expect User 1 to visit this location as other users like User 1 did not visit this location (or other locations like this one). We can use this surprise to flag this visit as a potential anomaly.

Despite the broad applications for anomaly detection in semantic trajectories, this area remains substantially under-researched. A main reason is a lack of semantic trajectory data with ground truth anomaly labels. Both privacy considerations and a lack of participant initiative result in datasets that are sparsely populated~\cite{Bonchianonymity,Suiprivacygroups,gao2017identifying}.
%
%
For example, the commonly used GeoLife dataset~\cite{geolife} includes only 182 users. But over the five years covered by the dataset, more than 60\% of the users have fewer than 50 recorded trajectories, and around 70\% of POIs are visited by only one or two users. In addition, this dataset has no ground truth on whether any segments of the trajectories can be considered as anomalous. 

We also step towards filling this research gap by providing trajectory datasets that have labels for anomalous trajectories. We provide (1) a dataset generated via user-based simulation for which we change the decision-making process of users while they are anomalous, and (2) a dataset generated from GeoLife in which we artificially introduce anomalous behavior by replacing the normal trajectory of   users with the trajectory of other users. 
%
%

To address the sparsity of the data, we leverage a deep learning approach to fill the gaps of missing data. Existing work in trajectory mining focusing primarily on temporal information for sequence embedding or integrating spatial information into Graph Neural Network (GNN) architectures~\cite{wang2020traffic,bui2022spatial}.
However, these approaches significantly increase computational costs~\cite{gnnchallenges} and require high-quality POI annotation. To address these challenges, we propose a collaborative filtering-based unsupervised anomaly detection approach in the context of human trajectory anomaly detection tasks.





To address those issues, we provide a modified neural collaborative filtering architecture to capture person-level human trajectory anomalous.
The main contributions of this work are as follows.

\begin{enumerate}[noitemsep,
        nosep,
        leftmargin=10pt,
        labelsep=2pt,
        itemindent=0pt]
  \item We present a neural collaborative filtering architecture to model the users' daily patterns of life without prior knowledge.
  \item We ascertain that neural collaborative filtering (NCF) can be effectively applied to the spatial-temporal anomaly detection domain. By employing an appropriate surprise function and training strategy, the model's cold start performance can be significantly enhanced.
  \item We perform extensive experiments on three highly sparse and extremely imbalanced datasets ~\cite{amiri2023massive,amiri2024patterns, geolife} to demonstrate the effectiveness and the promise of our neural collaborative filtering approaches on the user-level spatial-temporal anomaly detection domain.
\end{enumerate}

This paper is organized as follows:
In Section II, we explicate the basic concepts of user-level spatial-temporal trajectories and anomalous, as well as various anomaly detection methods and collaborative filtering techniques.
For better understanding, Section III introduces the fundamental concepts and definitions of our method.
%
%
Section IV introduces our methodology and its structure, detailing a Multi-Layer Perceptron (MLP) module and Matrix Factorization (MF) module. It also includes a discussion of the designed surprise functions and other relevant components of our approach.
Section V details the experiments conducted on several datasets using our methodology.
Finally, Section VI presents our conclusions and future work.

\vspace{-0.2cm}
\section{Related Works}
\vspace{-0.15cm}
\label{sec:related_works}
\begin{table*}[t]
    \centering
  
    \begin{adjustbox}{width=1.0\linewidth,center}
        \begin{tabular}{|c|c|c|c|c|c|}
            \hline
            UserId & $x_i$ (Latitude) & $y_i$ (Longitude)          & $tc_i$ (CheckinTime)         & $tl_i$ (LeavingTime)         & $s_i$ (VenueType)  \\
            \hline
            153    & 39.935892 & 116.453081 & 2009-06-21T03:03:01 & 2009-06-21T03:38:57 & Workplace  \\
            \hline
            153    & 39.998524 & 116.387211 & 2009-06-21T03:38:59 & 2009-06-21T04:08:00 & Restaurant \\
            \hline
            153    & 39.991694 & 116.389809 & 2009-06-21T04:27:56 & 2009-06-21T05:00:27 & Recreational        \\
            \hline
            153    & 39.991424 & 116.384395 & 2009-06-21T05:24:01 & 2009-06-21T05:53:31 & Recreational        \\
            \hline
            153    & 39.995926 & 116.390297 & 2009-06-21T06:12:41 & 2009-06-21T06:52:45 & Apartment  \\
            \hline
        \end{tabular}
    \end{adjustbox}
    \caption{ A template of human spatial-temporal trajectory datasets utilized in this paper. \andi{Having an example is very nice. But this example is very strange! This user goes to work at 3am and then travels to a restaurant in only two seconds. Also, let's remove the term Pub here and replace with ``Recreational Site''}\vspace{-0.5cm}}
    \label{table:spdataset}
    \vspace{-0.4cm}
\end{table*}

Here, 
we summarize widely used trajectory anomaly detection methods, both traditional and machine learning approaches in Section~\ref{sec:traj_outlier_related} followed by a introduction to the collaborative filtering method, covering its algorithm, applications, and recent developments in Section~\ref{sec:CF_related}.
\vspace{-0.2cm}
\subsection{Trajectory Anomaly Detection}\label{sec:traj_outlier_related}

Over the past few decades, a broad spectrum of trajectory anomaly detection methods has been introduced.
In this section, we systematically categorize these approaches into statistical, knowledge-driven, and machine learning approaches.
\vspace{-0.2cm}
\subsubsection{Statistical Approaches:} Those methods are commonly used in conventional anomaly detection. These methods rely on distance or probability distributions, such as the Gaussian distribution, to identify data points that significantly deviate from the norm.

Knorr and Ng ~\cite{knox1998algorithms} pioneered the concept of anomaly detection algorithms. In their distance-based approach, a data point is identified as an anomaly if fewer than $k$ other points are within a specified distance ($\delta$) from it in the dataset.
Liu et al.~\cite{liu2008isolation} introduced the Isolation Forest algorithm(IForest), which detects anomalies using binary trees. This method operates on the principle that anomalous are a minority and have attributes that differ markedly from normal instances. The algorithm isolates these anomalies by recursively partitioning the data into smaller subsets. If a forest of random trees consistently produces shorter path lengths for specific data points, those points are identified as anomalies.
%
Laxhammar et al. \cite{laxhammar2009anomaly} conducted a comparative study of the Gaussian Mixture Model (GMM) and Kernel Density Estimator (KDE) for anomaly detection in maritime traffic. Their study found that while KDE excels at capturing the intricacies of normal traffic, especially along sea lanes, both GMM and KDE have their limitations in detecting anomalies.
Ristic et al.\cite{ristic2008statistical} proposed a method to extract motion patterns from historical AIS data, which are then used to build anomaly detectors based on adaptive kernel density estimation (AKDE). This method supports real-time anomaly detection by sequentially applying the detector to incoming AIS data.
%
%
%
Wang et al.~\cite{wang2014trajectory} proposed a trajectory-based anomaly detection algorithm using multidimensional Hidden Markov Models (HMMs) for model training and likelihood estimation. Their approach allows for real-time detection of anomalous from sensor data, effectively identifying long-term deviations from normal behavior patterns.
Piciarell et al. ~\cite{piciarelli2008trajectory} developed a system for detecting anomalous events using trajectory analysis, particularly for video and traffic surveillance. Their system clusters anomalous trajectories with support vector machines (SVMs).
Thang \& Kim ~\cite{thang2011anomaly} proposed an improved DBSCAN algorithm, known as DBScan-MP, where each cluster may have different epsilon and minpts values, enhancing its adaptability and performance in identifying clusters and anomalous.
\vspace{-0.2cm}
\subsubsection{Knowledge-Driven Approaches} Knowledge-driven approaches leverage expert knowledge—such as traffic rules, navigational experience, and established physical models—to assess whether a ship, vehicle, or individual is exhibiting anomalous behavior \cite{brax2012self}. These approaches are particularly effective in domains like anomalous ship behavior detection (DBAS) and vehicle-to-vehicle (V2V) obstacle detection.

Avram et al. \cite{avram2012anomaly} introduced a method that combines expert knowledge with ship behavior data to develop a ship anomaly detector. This integrated approach optimizes purely data-driven models by refining their structure and making secondary corrections to their outputs. 
Wu et al. \cite{wu2017fast} developed the DB-TOD model to detect anomalous in vehicle trajectories. By capturing driving behavior and preferences from historical data, this probabilistic model calculates the latent costs of routing decisions using feature counts and latent biases, effectively identifying anomalous in both full and partial trajectories.
Zhu et al. \cite{zhu2015time} proposed the Time-Dependent Popular Route (TPRO) algorithm, which evaluates trajectory anomalies by considering spatial and temporal characteristics. The TPRO algorithm accounts for the varying popularity of routes between locations, assigning different weights to each when calculating anomaly scores.
Yuan et al. \cite{yuan2011trajectory} developed the Trajectory Outlier Detection based on Structural Similarity (TOD-SS) algorithm. This method segments trajectories based on angles at each point and calculates distances between sub-trajectories using features such as direction, speed, angle, and location. 
Yu et al. \cite{yu2018trajectory} presented the Trajectory Outlier Detection using Common Slice Subsequence (TODCSS) algorithm, which improves accuracy by combining features like direction, position, and continuity to detect anomalies across consecutive anomalous segments. 
Kinematics extraction, as discussed in studies like ~\cite{machado2019vessel,CuiKinematicVehicle,li2006motion}, Those methods integrates kinematic principles to produce precise and physically realistic trajectory predictions.
\subsubsection{Machine Learning Approaches} 
%
Recent advancements in machine learning have introduced promising solutions for anomaly detection, utilizing sophisticated neural network architectures to improve the identification of anomalies.
These developments offer enhanced capabilities for detecting complex patterns and deviations that traditional methods might miss.
Given the challenges of trajectory data sparsity and anomaly characteristics, Graph Neural Networks (GNNs) are increasingly employed in trajectory anomaly detection due to their ability to work with graph-structured data and ability to capture complex spatial patterns~\cite{sahili2023spatio}. 

Recently, there has been a significant advancement in GNN:
Goodge et al.~\cite{goodge2022lunar} proposed LUNAR which utilizes information from a node's nearest neighbors to identify anomalies.
Zhou et al.\cite{zhou2021ast} introduced AST-GNN, an attention-based spatio-temporal graph neural network that enhances pedestrian trajectory prediction by effectively capturing complex interactions and unique motion patterns.
Zheng et al.\cite{zheng2019addgraph} introduced AddGraph, which combines an extended temporal GCN with an attention-based GRU to detect anomalies in dynamic graphs. It captures node and edge information, updating node embeddings and hidden states at each time step. Anomaly scores are assigned to edges based on the associated nodes.

Another category of deep learning models, aside from GNNs, includes various neural network approaches that treat the trajectory as a series to learn complex patterns and identify anomalies known as series learning models.
These methods, including Recurrent Neural Networks (RNNs) and Long Short-Term Memory (LSTM) networks, are particularly effective for analyzing sequential or temporal data. 

For example: 
Niu et al.~\cite{niu2018alstm} introduce an adaptive LSTM that utilizes temporal continuity in sequential data. By incorporating a mask gate and maintaining span, this LSTM can update memory adaptively based on changes in the sequence input at each time step.
Autoencoder methods have also been effectively applied to trajectory anomaly detection. Liu et al. \cite{gmvsae} proposed a semi-supervised approach based on a Variational Autoencoder (VAE) called GM-VSAE, which models the probability distribution of route patterns in the latent space using a Gaussian mixture distribution. This method allows for the identification of anomalous by assessing the likelihood of a trajectory being generated from normal patterns.
Similarly, Han et al. \cite{deeptea} introduced DeepTEA, a semi-supervised method designed to detect time-dependent trajectory anomalous. DeepTEA employs a Convolutional Neural Network (CNN) to model road traffic conditions and utilizes a Gaussian Mixture VAE to capture latent trajectory patterns. 
\subsection{Collaborative Filtering}
\label{sec:CF_related}

Collaborative Filtering (CF), first introduced in the 1990s~\cite{10.1145/138859.138867}, is a widely used technique in recommendation systems. Originally, CF involved users collaborating to filter information by recording their reactions to documents they read, allowing others to access these reactions.
As a popular recommendation model, collaborative filtering utilizes ratings or other forms of user feedback to identify shared preferences among users. It then provides personalized recommendations based on these similarities, without requiring external information about the items or users.
Collaborative Filtering employs two main techniques: the neighborhood approach and latent factor models.
~\cite{korenAdvancesCollaborativeFiltering2022}.
The neighborhood approach models a user's preference for an item based on their ratings of similar items, while latent factor models utilize a shared latent factor space between items and users to explain ratings.
Nowadays, by leveraging advanced techniques such as matrix factorization ~\cite{bokde2015matrix}, singular value decomposition ~\cite{paterek2007improving} and integration of deep learning~\cite{wei2017collaborative,bobadilla2020deep,he2017neural} have further enhanced the accuracy and scalability of collaborative filtering models.

\vspace{-0.2cm}
\section{Preliminaries}
\label{sec:preliminary}

In this section, we present the key concepts and definitions essential to our study on trajectory anomaly detection used in the following sections.

\vspace{-0.2cm}
\subsection{Human Semantic Trajectory Representation}
A semantic trajectory is represented as a sequence of staypoints (or ``check-ins'') \( T = \{p_1 \rightarrow p_2 \rightarrow \dots \rightarrow p_n\} \) where $n$ is the number of staypoint observations in the trajectory.
Each staypoint observation \( p_i \) in this sequence is a six-tuple \( (userID, x_i, y_i, tc_i, tl_i, s_i) \), $userID$ is a unique identifier of an individual user, \( x_i \) is the longitude, \( y_i \) is the latitude, \( tc_i \) is the ``check-in'' timestamp when the trajectory \( T \) reaches point \( p_i (x_i, y_i) \), \( tl_i \) indicates the time the user leaves that point, and \( s_i \) provides semantic information about the visited location such as the type of PoI.
%
Table ~\ref{table:spdataset} provides an overview of the spatial-temporal trajectory for User 153. The first line of this example shows that this user stayed at a Workplace with the given coordinates from 03:03:01am (GMT) to 03:38:59am (GMT) and went to a restaurant shortly after.

The collection of all users' trajectories can be denoted as \( \mathcal{T} = \{T_1, T_2, \dots, T_m\} \), where \( m \) is the number of trajectories or users. 

Given the above definitions, in the next subsection, we formally formulate the trajectory anomaly detection problems.
\vspace{-0.2cm}
\subsection{Human Spatial-temporal Trajectory Anomaly Detection}
We assume the availability of (1) a training period during which we assume that users exhibit normal behavior and (2) a test period during which some users may exhibit anomalous behavior. The goal of trajectory anomaly detection is to identify the set of anomalous users during the test period using models of normal behavior learned using the train period. Formally,

\begin{definition}[Train and Test Period]
Let \( \mathcal{T} = \{T_1, T_2, \dots, T_m\} \) be a set of trajectories of $m$ users. We assume that any staypoint that started before a specific staypoint in time $t_{split}$ exhibits normal behavior and anomalous behavior may occur after $t_{split}$. Then, we can define a training trajectory:
$$
\mathcal{T}_i^{Train}=\{(userID, x_i, y_i, tc_i, tl_i, t_i)\in T_i| tc_i<=t_{split}\},
$$
and a testing trajectory:
$$
\mathcal{T}_i^{Test}=\{(userID, x_i, y_i, tc_i, tl_i, t_i)\in T_i| tc_i>t_{split}\}.
$$
This allows us to define a train dataset as 
$$
\mathcal{T}^{Train}=[T_1,...,T_m],
$$
and a test dataset as
$$
\mathcal{T}^{Test}=[T_1,...,T_m].
$$
\end{definition}

The goal of trajectory anomaly detection is to detect trajectories $T^{Test}_i\in T^{Test}$ what exhibit changes that deviate so much from their normal behavior that they rouse suspicion that they may have been generated by a different process. Intuitive, such different process may be different individuals (for example due to a stolen cellphone), a sudden change in behavior (such as exposure to an infectious disease), or a gradual change in behavior (such as the progress of a mental illness).

\begin{definition}[Trajectory Anomaly Detection]\phantom{HELLO!} \newline
Let \( \mathcal{T} = \{T_1, T_2, \dots, T_m\} \) be a trajectory dataset split into $T^{Train}$ and $T^{Test}$, and let $P\subseteq T^{Test}$ be a set a anomalous trajectories. The task of Trajectory Anomaly detection is to find $P$. 
\end{definition}

Given a training set of users \( \mathcal{U} \) and their normal trajectory pairs \( \mathcal{T}_u \), the objective is to identify anomaly users \( U_{anomaly} \) in the test set who exhibit substantial deviations from the typical trajectory patterns established across the training population. These deviations are quantified using a score function \( f_{score} \), which measures the extent of divergence from the common patterns observed within \( \mathcal{U} \).

This goal presents several unique challenges:
\begin{enumerate}[noitemsep,
        nosep,
        leftmargin=10pt,
        labelsep=2pt,
        itemindent=0pt]
    \item \textbf{Challenges in sparse data:} Accurately clustering users based on their patterns of life involves addressing complex issues related to both inter-user interactions and the broader context of their trajectories.
One challenge as we mentioned before is the reliable formation of user clusters from datasets that are both sparse and consist of only positive pairs.
The challenge lies in ensuring that the model consistently generates accurate user pattern clusters from a dataset that is only include positive pairs and highly sparse.
Many POIs are visited by only a few users, which exacerbates the difficulty of achieving comprehensive clustering.
To address this, the clustering algorithm must effectively identify and interpret meaningful interactions among users despite the limited data coverage.
Meanwhile, these methods must be robust enough to ensure prediction accuracy, even when data is unevenly distributed.
\item \textbf{Challenges in quantifying anomalous:} Human behavior is inherently dynamic and can be influenced by various temporal and spatial factors.
Traditional anomaly detection methods may be inadequate in this context, as they might misclassify new or infrequent POIs as anomalies.
This limitation underscores the need for more sophisticated approaches to accurately quantify anomalous.
Sophisticated approaches are required to accurately identify and score anomalous, effectively distinguishing between genuine anomalies and legitimate but less frequent behaviors.

\end{enumerate}
%

\section{Methodology}
\label{sec:methodology}

\begin{figure*}[t]
    \centering
    \includegraphics[width=.85\linewidth,trim=0cm 1.1cm 0.3cm 0.96cm,clip]{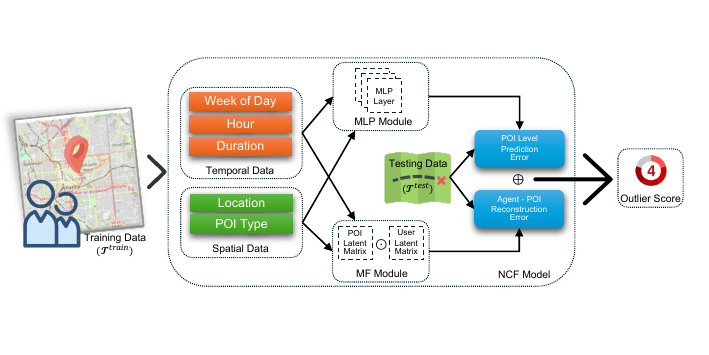}
    \vspace{-0.4cm}
    \captionsetup{width=1.\linewidth}
    \caption{The illustration of our proposed NCF model. (Left) Extraction of spatio-temporal embeddings from input trajectories;  (Middle) Neural Collaborative Filtering with two-level of prediction; (Right) Anomaly scores based on weighted prediction.\vspace{-0.45cm}}
    \label{fig:framework}
    \vspace{-0.1cm}
\end{figure*}

In this section, we introduce our neural collaborative filtering-based unsupervised anomaly detection approach for human trajectory analysis. While this section provides ideas, concepts, and theory, the interested reader may find implementation details in our GitHub Repository at~\url{https://github.com/alex-cse/NCF_AHSTD}. 
Notably, our method can identify anomalous users without any explicit semantic information of places. Instead, our approach extracts latent semantic information using collaborative filtering (based on users visiting similar places and places visited by similar users).
Our framework comprises two modules:
\begin{enumerate}[noitemsep,
        nosep,
        leftmargin=10pt,
        labelsep=2pt,
        itemindent=0pt]
    \item The Collaborative Filtering Module uses collaborative filtering to model the User-POI visit matrix observed in the train period. It identifies anomalous users by measuring surprise, that is by detecting deviations from typical (model) trajectory patterns observed in the test period.  
    \item Neural Module: This module take the advantage of neural networks to interpret the complex spatial-temporal relationships within the data. Traditional methods like collaborative filtering often fall short when dealing with those spatial-temporal data because they may not fully capture non-linear patterns or subtle dependencies within the dataset.
\end{enumerate}
To effectively quantify anomalous users, we employ an accumulator based on a continuous-state Markov chain to calculate anomaly scores.
This approach is grounded in the principle that each user's trajectory is conceptualized as a sequence where each event is influenced by the preceding one, but making the simplifying (Markov-)assumption that the future is conditionally independent of the past given the present. 
We assume that when a trajectory exhibits anomalous behavior, this behavior not only affects a single staypoint, but affects a chain of subsequent events. For instance, this might manifest as an increase in travel distance or as visits to a POI during an atypical time frame.
To capture these anomalies, the accumulator evaluates each POI for every user based on the trajectory sequence.
By doing so, our method can detect not only extreme anomalous but also users with moderate deviations from the expected patterns in their overall trajectory.
%
%
%
\subsection{Neural Collaborative Filtering Model}\label{section:model}
\vspace{-0.1cm}
To thoroughly explore and understand the spatio-temporal interactions between users and POIs, we propose a streamlined neural collaborative filtering approach~\cite{rendleNeuralCollaborativeFiltering2020,he2017neural}.
This model leverages a Multi-Layer Perceptron (MLP) module to effectively capture and learn both the temporal and spatial mobility of users, as well as the spatio-temporal relationships among various locations.
By integrating the MLP module with a matrix factorization framework, we enable a comprehensive analysis of users' behaviors over time and the intricate spatial interactions between different POIs.

Our proposed architecture is illustrated in Figure \ref{fig:framework}.
Specifically, within the trajectory pairs denoted as \(\mathcal{T}\), we analyze the interactions between users \(u\) and their respective POI \(p_{i}\), forming trajectory pairs \(\mathcal{T}_{ui}\).
In the context of \(\mathcal{T}_{ui}\), we consider several key parameters.\vspace{-0.1cm}
\begin{itemize}
    \item \(h \in [0, 23]\) is the hour of check-in, providing a temporal resolution on an hourly basis.
    \item \(d \in [0, 6]\) is the day of the week when the check-in occurs.
    \item The geographical location \(l\) specifies the geographical location of the check-in, which is critical for spatial analysis.
\end{itemize}
Additionally, we preprocess the data within each trajectory pair \(\mathcal{T}_{ui}\) make it encompassed a set of movement attributes, including stay duration, and distance traveled.
The stay duration indicates the time spent at a particular location, while movement speed and distance provide insights into the mobility patterns and travel behaviors of the user.
The final tuples $\mathbf{Z}$ of user $i$ and POI $j$ pair can be represent as:
\begin{equation}
    \mathbf{Z_{i,j}}=\begin{bmatrix}
        u_i \\
        h_j \\
        d_j \\
        l_j \\
    \end{bmatrix}
\end{equation}
In this context, \(h_j\), \(d_j\), and \(l_j\) denote the embeddings corresponding to the hour of the user's arrival at the POI, the week of of day the user's visit the POI, and the movement distance from previous check-in POI (calculate base on latitude and longitude) of the POI, respectively. 
After that we applied a three-layer MLP module to learn the interactions of latent features, the equation can represented as:
\begin{equation}
    \phi^{MLP} = a_3 \left( W_3^T \left( a_2 \left( W_2^T \left( a_1 \left( W_1^T \mathbf{Z} + \mathbf{b}_1 \right) \right) + \mathbf{b}_2 \right) \right) + \mathbf{b}_3 \right)
\end{equation}
Here,  \(\mathbf{Z}\) is the input latent feature vector,  $W_i$  and  $\mathbf{b}_i$ are the weight matrices and bias vectors of each layer, and  $a_i$  represents the activation functions applied at each layer.
For the Generalized Matrix Factorization component, we also utilize embeddings derived from the vectors \(\mathbf{Z}\), incorporating additional information about the POI type \(t\).
Given that users \( u_i \) have features like hour \( h_i \), week of day \( w_i \), and distance \( d_i \), and POI \( p_j \) have features like type \( t_j \) , the equation for matrix factorization incorporating these auxiliary features can be written as:
\begin{equation}
     \phi^{MF}_{ij} \approx u_i^T p_j + (w_1 \cdot h_i + w_2 \cdot d_i + w_3 \cdot h_i) + (w_4 \cdot t_j)
\end{equation}
Where the \( w_1, w_2, w_3, w_4 \) are weights for the corresponding auxiliary features.
This equation combines the latent factor model with additional terms that account for the auxiliary features, allowing us to capture both latent features (inferred from similar users visiting the same POI) and explicit features (such as known types of POIs like restaurants or schools).

In our Spatial-Temporal model, we extend beyond the conventional method by integrating not only user and POIs latent variables but also additional latent features specifically designed to capture spatial and temporal dynamics.
This includes geographic information that captures the spatial context of the interactions, as well as temporal attributes that detail the duration of stays and specific visit times.
The geographic information provides a framework for understanding the spatial relationships and locations of the staypoints of interest, while the temporal attributes offer insights into the timing and frequency of visits.
After the initial embedding layer, the model utilizes multiple fully connected layers, which function as neural collaborative filtering layers.
%
The overall MLP module was trained on the training dataset \( \mathcal{T}^{Train} \), with the objective of capturing the representation of normal patterns.
Once trained, any deviations from these established patterns are indicative of anomalies in the test dataset \( \mathcal{T}^{Test} \). In the following section, we describe how to measure these deviations and to identify anomalous users.
%
%
%
\vspace{-0.3cm}
\subsection{Quantification of Anomaly Scores} \label{section:regular}

We propose two measures to quantify the degree to which an individual is anomalous: The first measure directly considers the difference (the surprise) between the expected (by the NMF model) score $\phi^{MF}_{ij}$  of an individual $i$ visit a place $j$ (described in the previous section) and the observed visits $\mathcal{T}^{Test}$. While some places are visited by a large number of individuals every day, other places may have very few or no daily visits. For such places, it becomes difficult to predict which individual may visit them, and, consequently, a (normal) visit may incur a large surprise to the model. To avoid incurring such false positives, we additionally propose not only to predict (and measure surprise) for individual POIs, but also for the type of POI visited. As an intuitive example, it may surprise the model that individual $A$ visits, during lunch time, a new German restaurant that none have visited during the training period (because the restaurant may not have opened yet). But it is not surprising that $A$ visits a restaurant during lunchtime as they always do. This section describes details for both these measures.
\vspace{-0.3cm}
\subsubsection{User-POI Matrix Reconstruction Surprise:}\label{subsubsec:surprise1} 
This surprise function assesses the discrepancy between the User-POI visit status matrix in the test set and its training set Matrix Factorization reconstruction.
The User-POI visit status matrix is a binary matrix where each entry signifies whether a specific user has visited a particular POI.
A high difference in the User-POI Reconstruction Error indicates that the reconstructed matrix significantly deviates from the actual visit status matrix, suggesting that the user’s behavior is inconsistent with the expected patterns and should be considered anomalous.
%
%
In the training matrix, each observed User-POI pair is given a value of 1 if the corresponding user visited the corresponding POI, and 0 otherwise. The factorized model aims at reconstructing this matrix using compressed information, thus assigning a visit score to each User-POI pair. We note that these scores may be less than 0 or greater than 1 and thus, should not be interpreted as probabilities.\footnote{Although we note that using a logistic model that maps scores to probabilities in the interval [0,1] may be an interesting extension of the model.}

Applying a Generalized Matrix Factorization provides a matrix $\phi^{MF}$ with dimensions \(m \times n\), where \(n\) represents the number of individuals and \(m\) denotes the number of POIs. We propose to simply use the absolute difference between this prediction and the observed visits as follows:
\begin{equation}\label{eq:surprise}
    \text{surprise}_{ij} = \left| \phi^{MF}_{ij} -\mathcal{T}^{Test}_{ij}\right|
\end{equation}
This method ensures that user-POI visit will receive a low surprise if the user visits the POI in the test period but the POI is considered a typical location based on the collaborative filtering. Conversely, an anomalous POI, which deviates from the expected pattern, will be assigned a higher surprise. We note that this surprise function also gives a high surprise to a place that is NOT visited in the test period but is expected to be visited by the collaborative filtering model. 

If we have additional information about the semantics of the anomalies that were expecting to find, we can adopt the above surprise function accordingly. For example, if we know that anomalous agents visit places they would not normally visit (such as going to random places), we can reflect this apriori knowledge in the surprise by only counting surprise for POIs visited in Test:
\begin{equation}
    \text{surprise}^{\text{NewPOI}}=  \min (0,\mathcal{T}^{Test}_{ij}-\phi^{MF}_{ij})
\end{equation}
Analogously, if we know that anomalous agents stop visiting places they would normally visit (such as skipping going to work), we can adopt the surprise function to only consider POIs that were not visited in Test: 
\begin{equation}\label{eq:surprise3}
    \text{surprise}^{\text{MissingPOI}}=  \min (0,\phi^{MF}_{ij}-\mathcal{T}^{Test}_{ij})
\end{equation}
If we have no apriori knowledge about how anomalies manifest, we can use the surprise function $\text{surprise}_{ij}$ in Equation~\ref{eq:surprise}.

The surprise functions defined in Equations~\ref{eq:surprise}-\ref{eq:surprise3} provide us with anomaly scores for individual individual-POI pairs. To obtain anomaly scores at the individual user level, we aggregate these scores by individual. 
The accumulated error for user $i$ can be computed as:
\begin{equation}
    \text{surprise}_{i} = \sum_{j=1}^{m} \text{surprise}_{ij}
\end{equation}
The implicit assumption that we make here is that an anomalous user may have more than a single anomalous visit. Depending on apriori information of what may constitute anomalous behavior of an individual user, we can adapt this surprise function accordingly. For instance, if we know that an anomalous user will visit exactly one anomalous place, then we can adjust this user-level surprise function by computing the maximum surprise over all POIs of the same user:
\begin{equation}
    \text{surprise}^{\text{singlePOI}}_{i} = \max_{j=1}^{m} \text{surprise}_{ij}
\end{equation}

%

\subsubsection{POI Type Surprise}\label{subsubsec:surprise2}
In addition to measuring surprise at individual POI level, we also propose to measure surprise the the POI type level. 
The surprise function at the POI type level quantifies the divergence between the model’s predicted POI type and the actual POI type. This approach is specifically designed to address the issue of data sparsity: Even though individual POIs may not have enough visits to allow the collaborative filtering to model the latent features of the POI.

Given a set \( C=\{c_1,...,c_{|C|}\} \) of types (classes) of POIs, and letting $c(p_j)$ be the class of POI $p_j$, then the probability of class \( k\in C \) visited by user $u_i$ can be calculated as:
\begin{equation}
    p_{ik} = 
\frac{\sum_{j=1}^{m}\mathcal{T}^{train}_{ij}\cdot \mathbb{I}(c(p_j)=k)}{\sum_{j=1}^{m}\mathcal{T}^{train}_{ij}}
\end{equation}
where $m$ is the number of POIs and $\mathbb{I}(c(p_j)=k)$ is an indicator function that returns $1$ if POI $p_j$ has type $k$ and zero otherwise.

Then, for each user we can define their most likely POI type, as:
$$
\text{mostLikelyPOI}_i=\arg\max{k\in C}(p_{ik})
$$

To measure surprise of a user $u_i$, we count the number of visits observed during the test period that do not match the user's most likely POI:

$$
\text{surprise}^{\text{POIType}}_i= \sum_{j=1}^{m} \mathbb{I}(\text{mostLikelyPOI}_i \neq \mathcal{T}_{Test}(i,j)) \cdot \mathcal{T}_{Test}(i,j)
$$

%
Our final anomaly score of each user is then defined as the sum of the two surprise scores:
\begin{equation}
    \text{AnomalyScore}_{i} =  \text{surprise}_{i} + \text{surprise}^{\text{POIType}}_i
\end{equation}
where $\text{surprise}_{i}$ is the User-POI Matrix Surprise described in Section~\ref{subsubsec:surprise1} and $\text{surprise}^{\text{POIType}}_i$ is the POI Type Surprise described in Section~\ref{subsubsec:surprise2}.

\section{Experimental Results and Environmental setup}
\label{sec:results}

\begin{table*}[t]

    \centering
    \begin{adjustbox}{width=.96\textwidth,center}
        \begin{tabular}{@{}lcccccccc@{}}
            \toprule
                     & \multicolumn{4}{c}{ATL}     & \multicolumn{4}{c}{NOLA}                                                                                                               \\
            \cmidrule(lr){2-5} \cmidrule(l){6-9}
            Model    & Top-10 Hits                 & Top-100 Hits             & Top-150 Hits  & AUC score          & Top-10 Hits   & Top-100 Hits       & Top-150 Hits & AUC score          \\
            \midrule
            OMPAD    & 0                           & 4                        & 5            & 0.4782             & 1             & 1                 & 1           & 0.4509             \\
            DSVDD    & \textbf{10}               & \underline{28}              & \underline{29}            & \underline{0.6171}             & \textbf{9}    & 17                 & \underline{26}           & 0.6098             \\
            DAE      & 1                           & 6                       & 7            & 0.4759             & 0             & 3                  & 6           & 0.4937             \\
            ECOD  & 3        & 12                       & 15            & 0.5357             & 4             & \underline{18}                 & 23           & \underline{0.6274}             \\
            IForest  & 2        & 6                       & 11            & 0.5136             & 0             & 5                 & 13           & 0.5247             \\
            
            \midrule
            NCF      & \underline{9}                  & \textbf{35}           & \textbf{43}            & \textbf{0.6853}    & \underline{7} & \textbf{44}        & \textbf{58}           & \textbf{0.7760}    \\
            \midrule

                     & \multicolumn{4}{c}{FVA}     & \multicolumn{4}{c}{BJNG}                                                                                                               \\
            \cmidrule(lr){2-5} \cmidrule(l){6-9}
            OMPAD    & 1                           & 1                        & 1            & 0.4796             & 1             & 5                  & 6           & 0.4370             \\
            DSVDD    & \underline{5}               & \underline{19}                       & \underline{24}           & \underline{0.6329}             & \textbf{10}   & \underline{29}     & \underline{33}          & \underline{0.5643}            \\
            DAE      & 3                           & 12                        & 17            & 0.5404             & 1             & 7                 & 14           & 0.5474             \\
            ECOD  & 1                           & 12                       & 14            & 0.5318             & 1             & 17                 & 23           & 0.5778             \\
            IForest  & 2        & 7                       & 13            & 0.5616             & 0             & 7                 & 13           & 0.5274             \\

            \midrule
            NCF      & \textbf{7}                  & \textbf{33}              & \textbf{41}            & \textbf{0.6673}    & \underline{9} & \textbf{42}        & \textbf{48}           & \textbf{0.7209}    \\
            \midrule
                     & \multicolumn{7}{c}{GeoLife}                                                                                                                                          \\
            \cmidrule(lr){3-7}
                     &                             & OMPAD                    & 0             & 0                  & 1            & 0.0552                                                 \\
                     &                             & DSVDD                    & \textbf{4}    & \textbf{9}        & \textbf{16}            & \textbf{0.8714}                                        \\
                     &                             & DAE                      & \textbf{4}             & 4                 & 7            & 0.4422                                                 \\
                     &                            & ECOD                  & \textbf{4}             & \underline{6}                 & \underline{11}            & \underline{0.6977}                                                 \\
                     &                             & IForest                  & \underline{3}             & 5                 & \underline{11}            & 0.6531                                                 \\            
            \midrule
                     &                             & NCF                      & \underline{3}             & \underline{6}                  & 7            & 0.6236                                                 \\

            \bottomrule
        \end{tabular}
    \end{adjustbox}
    \caption{Anomaly detection performance. For each dataset, the highest AUC scores are highlighted in bold, and the second-highest scores are underlined. We report Top-5/10/25 Hits for GeoLife dataset due to its size constraint.  \vspace{-0.2cm}}\label{table:performance}
  
\end{table*}
In this section, we conduct a comprehensive evaluation of our neural collaborative filtering model using multiple simulated and real-world datasets. The subsequent parts of this section will provide detailed information about each dataset, including their anomaly characteristics and the specific preprocessing steps applied. Additionally, we will outline the anomalous settings and configurations in the simulated datasets used for our model.
%
%
resent a thorough evaluation of our model's performance.
%
%
%
%
\vspace{-0.3cm}
\subsection{Experimental Datasets}
\label{sec:dataset}

To evaluate the performance of our proposed model, we utilized simulated and real-world dataset. The simulated datasets were generated using the Agent-Based Patterns-of-Life Simulation (POL)~\cite{zufle2023urban,kim2020location}, while the real-world dataset was based on the GeoLife dataset~\cite{zheng2010geolife,zhang2023large,zhang2024transferable}. 


\vspace{-0.2cm}
\subsubsection{Agent-Based Simulation of Patterns of Life}
The Patterns of Life simulation is an existing simulation that was designed to emulate human needs and behaviors in an urban environment~\cite{zufle2023urban,kohn2023epipol,amiri2024urban}. Within the simulation, agents engage in activities that mimics real-life actions, such as attending work, forming friendships, and participating in social gatherings. These agents navigate a virtual environment modeled on real-world settings like roads and buildings, sourced from OpenStreetMap.\\ 
In our study, we test our model on four distinct map setting: Fairfax County, Virginia (FVA); the French Quarter of New Orleans, Louisiana (NOLA); Atlanta, Georgia (ATL); and Beijing, China (BJNG). 
These dataset contains 450 days of normal life, followed by an additional 14 days with introduce anomalous behaviors into the regular patterns which serve as the training and test dateset.
In the 14-day test dataset, we identified three distinct types of anomalous behaviors: hunger anomalous, social anomalous, and work anomalous. Those anomalous behaviors are defined as follows:
\begin{itemize}[noitemsep,
        nosep,
        leftmargin=10pt,
        labelsep=2pt,
        itemindent=0pt]
    \item Hunger anomalous: An agent under this category becomes hungry more quickly. Such agents have to go to restaurants or their homes much more often to satisfy their food needs.
    \item Social anomalous: This type of agent randomly selects recreational sites to meet and spend leisure time, rather than being guided by their attributes and social network.
    \item Work anomalous: Agents in this category do not go to work when they normally should be working.
\end{itemize}


These anomalies were further classified into three intensity levels: red (high intensity), orange (mid intensity), and yellow (low intensity).
The detailed intensity levels are defined as follows:
\begin{itemize}[noitemsep,
        nosep,
        leftmargin=10pt,
        labelsep=2pt,
        itemindent=0pt]
    \item Yellow Anomalies: The agent displays anomalous behavior at a low intensity. For example, a hunger anomaly with a yellow intensity level experiences hunger more frequently than usual but does not need to visit a restaurant or home as often as a red intensity anomaly. For work and social anomalous, the yellow intensity level indicates that the agents will behave anomalously 20\% of the time.
    \item Orange Anomalies: The agent displays anomalous behavior at a moderate intensity. For example, a hunger anomaly with an orange intensity level feels hungry more frequently than normal and needs to visit a restaurant or home more often than a yellow intensity anomaly, but less often than a red intensity anomaly. For work and social anomalous, the orange intensity level indicates that the agents will behave anomalously 50\% of the time.
    \item Red Anomalies: The agent displays anomalous behavior at the highest intensity. For example, a hunger anomaly with a red intensity level experiences hunger more intensely than other anomalous and needs to visit a restaurant or home more frequently. For work and social anomalous, the red intensity level indicates that the agents will behave anomalously 100\% of the time.
\end{itemize}

\vspace{-0.2cm}
\subsubsection{Real World Dataset}\label{app:geolife}
The real-world dataset for this study was derived from the Microsoft Research Asia's GPS Trajectory dataset~\cite{zheng2010geolife}. We employed an stay-point extraction algorithm~\cite{zheng2008mining} to transform the data into a check-in pattern suitable for our studies. Subsequently, we utilized OpenStreetMap to categorize locations into four groups: apartments, workplaces, recreational, and restaurants. Given the extensive range of categories in OpenStreetMap, we manually classified them into these distinct groups.
Following preprocessing, agents with fewer than 50 records were excluded, yielding a final dataset consisting of 69 agents. This refined dataset includes 14,080 training trajectories and 3,552 test trajectories, spanning a period of over four years. Within this dataset, we introduced a specific type of anomaly known as the "imposter anomaly," where an agent switches trajectories with another agent after a certain point in time. The dataset was divided into two segments: 80\% for training, using the stay points, and the remaining 20\% for testing, where anomalous were introduced.

\vspace{-0.3cm}
\subsection{Experimental Settings}

In this section, we introduce our competitor methods and evaluation metrics.
\vspace{-0.1cm}
\subsubsection{Competitor Methods:\label{appendix:comarison}}
We compare with several trajectory anomaly detection methods, including non-deep learning methods and learning methods:\\
\textbf{OMPAD} \cite{basharat2008learning} is an anomaly detection method that analyzes objects' movement patterns by counting the types of locations they visit. It identifies abnormal activities by measuring the deviations in moving trends compared to established normal patterns.\\
\textbf{DSVDD} \cite{ruff2018deep} is a deep one-class classification based anomaly detection method. We generalize it to handle the task of semantic trajectory anomaly detection in a most intuitive way. We map the weekly trajectories of each user to a high dimensional sphere by a deep neural network encoder. Then the distance of trajectories from the sphere's surface is quantified as an anomaly score.\\
\textbf{DAE} \cite{zhou2017anomaly,dotti2020hierarchical} is a widely-used anomaly detection method that leverages a deep autoencoder. Utilizing an encoder-decoder model architecture, it reconstructs input trajectories, and the resulting reconstruction error is used as an anomaly indicator, signifying deviations from the normal pattern.\\
\textbf{ECOD} \cite{Li_2023} is an unsupervised anomaly detection algorithm that uses empirical cumulative distribution functions to identify anomalous without requiring parameter tuning, offering both efficiency and interpretability.\\
\textbf{IForest} \cite{liu2008isolation} is an algorithm detects anomalies by recursively partitioning data, isolating anomalous based on shorter path lengths in a forest of random trees.
\subsubsection{Evaluation Metrics}
To rigorously evaluate the performance of our anomaly detection methodology, we utilize the Top-K hits metric.
In our approach, users with the K highest anomaly scores are designated as anomalous.
The number of hits serves as an indicator of the method's efficacy in accurately distinguishing anomalous.
The Top-K hits of the model refer to the true positives, representing the anomaly users correctly identified within the top K ranked predictions made by the model.
For our evaluation, we specifically use Top-10, Top-100, and Top-150 hits. However, for the GeoLife dataset, we report Top-5/10/25 hits instead of Top-100 due to anomaly number constraints.
Furthermore, we employ the area under the receiver operating characteristic curve (AUC) as an additional evaluation metric.

\begin{table*}[t]
    \centering
    \begin{adjustbox}{width=.80\textwidth,center}
        \begin{tabular}{@{}lcccccccccccc@{}}
            \toprule
            & \multicolumn{3}{c}{ATL} & \multicolumn{3}{c}{NOLA} & \multicolumn{3}{c}{FVA} & \multicolumn{3}{c}{BJNG} \\
            \cmidrule(lr){2-4} \cmidrule(lr){5-7} \cmidrule(lr){8-10} \cmidrule(lr){11-13}
            Category & Red & Orange & Yellow & Red & Orange & Yellow & Red & Orange & Yellow & Red & Orange & Yellow\\
            \midrule
            Hunger & 0.30 &0.23 &0.30 &0.40 &0.53 &0.33 &0.36 &0.30 &0.30 &0.40 &0.43 &0.43 \\
            Work   & 0.40 &0.30 & 0.10 &0.70 & 0.40  &0.30  & 0.50  &0.20 & 0.10 &0.50 & 0 &0.20  \\
            Social & 0.60  &0.40 & 0 &0.50   & 0.10  &0  & 0.30  &0.10 & 0 &0.30  & 0  &0  \\
            \bottomrule
        \end{tabular}
    \end{adjustbox}
    \caption{Recall of each category in the top 150 predictions of each dataset. (Recall@150)\vspace{-0.4cm}}
    \label{table:categorycount}
    \vspace{-0.5cm}
\end{table*}

\vspace{-0.3cm}
\subsection{Anomaly Detection Results}
\textbf{Detection Results:} Table~\ref{table:performance} presents the anomaly detection performance of our proposed method alongside several competitive methods. In the table, the highest AUC scores for each dataset are highlighted in bold, while the second-highest scores are underlined.
The results show for the four simulated datasets, our NCF model always achieves the highest AUC scores by a substantial margin compared to other machine learning and non-machine learning methods. However, for the GeoLife real-world dataset, our proposed method is significantly outperformed by state-of-the-art methods. We explain this by the extreme sparsity of the GeoLife dataset. This dataset captures only 182 across all of Beijing, China. Having so few users, there are very few cases where the same place is visited by more than one unique user. Thus, the resulting User-POI matrix only has one non-zero value in most columns. This makes it impossible for the matrix factorization to successfully apply any collaborative filtering. Thus, Table~\ref{table:performance} shows promising results in the case where we have more dense data (for all users and all places). But for extremely small datasets like GeoLife, collaborative filtering is not able to infer any common topics and features among the users and places due to a lack of examples where different users visit the same place.

\begin{figure}[t]
    \centering
    \includegraphics[width=0.99\linewidth,trim=0cm 0cm 0.1cm 0cm,clip]{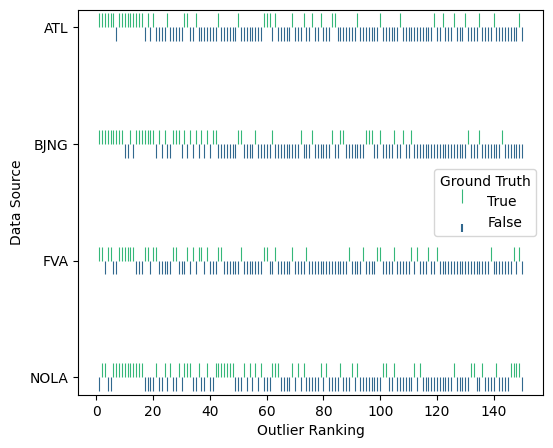}
    \vspace{-0.6cm}
    \captionsetup{width=1.\linewidth}
    \caption{The anomaly ranking among four datasets\vspace{-0.4cm}}
    \label{fig:ranking}
    \vspace{-0.3cm}
\end{figure}

To give more insights on the quality of the anomaly scores returned by NCF, beyond a single AUC-PR score, Figure~\ref{fig:ranking} shows how true positives and true negatives are distributed across the anomaly score ranking. We observe a clear correlation that true positives are frequently among the highest ranked anomaly scores.\\
%
%
\textbf{Detailed Detection Count for Types of anomalous:} To evaluate the detection performance for each anomaly category Table~\ref{table:categorycount} provides the true positive detection count per category in the top 150 predictions of each dataset.
This result indicates that, in most cases, our model consistently achieves high detection accuracy for anomalous with the highest intensity, particularly demonstrating heightened sensitivity to categories involving significant spatial changes.
%
This observation indicates that our method is particularly sensitive to variations in spatial patterns but less responsive to temporal patterns. 

\vspace{-0.2cm}
\section{Conclusions and Future Work}
\label{sec:conclusion}

In conclusion, this paper presents an unsupervised method for detecting anomalous in human trajectories using a neural collaborative filtering model applied to user-level spatiotemporal datasets. 
Our approach effectively addresses challenges such as data sparsity and imbalance by leveraging the strengths of the collaborative filtering module and neural network components, which significantly mitigate these issues. 
Comprehensive experiments on our five datasets validate the effectiveness and efficiency of the proposed method, which shows substantial improvements in accurately detecting anomalous within complex spatiotemporal datasets. 
For future research, we aim to explore the impact of geolocation embedding factors on model performance across various types of anomalies. 
Additionally, we will investigate incorporating more noise data to enhance the robustness of anomaly detection in real-world scenarios.

\vspace{-0.4cm}

\section{Acknowledgements}
Supported by the Intelligence Advanced Research Projects Activity (IARPA) via Department of Interior/ Interior Business Center (DOI/IBC) contract number 140D0423C0025. The U.S. Government is authorized to reproduce and distribute reprints for Governmental purposes notwithstanding any copyright annotation thereon. Disclaimer: The views and conclusions contained herein are those of the authors and should not be interpreted as necessarily representing the official policies or endorsements, either expressed or implied, of IARPA, DOI/IBC, or the U.S. Government.

\vspace{-0.34cm}

\bibliographystyle{ACM-Reference-Format}
\bibliography{refs/main}

\end{document}